# Learning to Color from Language

Varun Manjunatha★♠○  Mohit Iyyer★†‡  Jordan Boyd-Graber♠♣○◇  Larry Davis♠○

University of Maryland : Computer Science♠, Language Science◇, iSchool♣, UMIACS○
Allen Institute of Artificial Intelligence†  University of Massachusetts, Amherst‡
{varunm@cs, jbg@umiacs, lsd@umiacs}.umd.edu  miyyer@cs.umass.edu

## Abstract

Automatic colorization is the process of adding color to greyscale images. We condition this process on language, allowing end users to manipulate a colorized image by feeding in different captions. We present two different architectures for language-conditioned colorization, both of which produce more accurate and plausible colorizations than a language-agnostic version. Through this language-based framework, we can dramatically alter colorizations by manipulating descriptive color words in captions.

## 1 Introduction

Automatic image colorization (Cheng et al., 2015; Larsson et al., 2016; Zhang et al., 2016; Iizuka et al., 2016; Deshpande et al., 2017)—the process of adding color to a greyscale image—is inherently underspecified. Unlike background scenery such as sky or grass, many common foreground objects could plausibly be of any color, such as a person's clothing, a bird's feathers, or the exterior of a car. Interactive colorization seeks human input, usually in the form of clicks or strokes on the image with a selected color, to reduce these ambiguities (Levin et al., 2004; Huang et al., 2005; Endo et al., 2016; Zhang et al., 2017). We introduce the task of colorization from natural language, a previously unexplored source of color specifications.

Many use cases for automatic colorization involve images paired with language. For example, comic book artwork is normally first sketched in black-and-white by a penciller; afterwards, a colorist selects a palette that thematically reinforces the written script to produce the final colorized art. Similarly, older black-and-white films are often colorized for modern audiences based on cues from dialogue and narration (Van Camp, 1995).

Language is a weaker source of supervision for colorization than user clicks. In particular, language lacks ground-truth information about the colored image (e.g., the exact color of a pixel or region). Given a description like *a blue motorcycle parked next to a fleet of sedans*, an automatic colorization system must first localize the motorcycle within the image before deciding on a context-appropriate shade of blue to color it with. The challenge grows with abstract language: a red color palette likely suits an artistic rendering of *the boy threw down his toy in a rage* better than it does *the boy lovingly hugged his toy*.

We present two neural architectures for language-based colorization that augment an existing fully-convolutional model (Zhang et al., 2016) with representations learned from image captions. As a sanity check, both architectures outperform a language-agnostic model on an accuracy-based colorization metric. However, we are more interested in whether modifications to the caption properly manifest themselves in output colorizations (e.g., switching one color with another); crowdsourced evaluations confirm that our models properly localize and color objects based on captions (Figure 1).

## 2 Model

This section provides a quick introduction to color spaces (Sec. 2.1) and then describes our baseline colorization network (Sec. 2.2) alongside two models (Sec. 2.3) that colorize their output on representations learned from language.

### 2.1 Images and color spaces

An image is usually represented as a three dimensional tensor with red, green and blue (RGB) channels. Each pixel's color and intensity (i.e., lightness) are *jointly* represented by the values of these three channels. However, in applications such as colorization, it is more convenient to use representations that separately encode lightness and color. These *color spaces* can be obtained through mathematical transformations of the RGB color space; in this work, following Zhang et al. (2016), we use the CIE *Lab* space (Smith and Guild, 1931). Here, the first channel (L) encodes only lightness (i.e., black-and-white). The two color channels $a$ and $b$ represent color values between green to red and blue to yellow, respectively. In this formulation, the task of colorization is equivalent to taking the lightness channel of an image as input and predicting the two missing color channels.

★Authors contributed equally

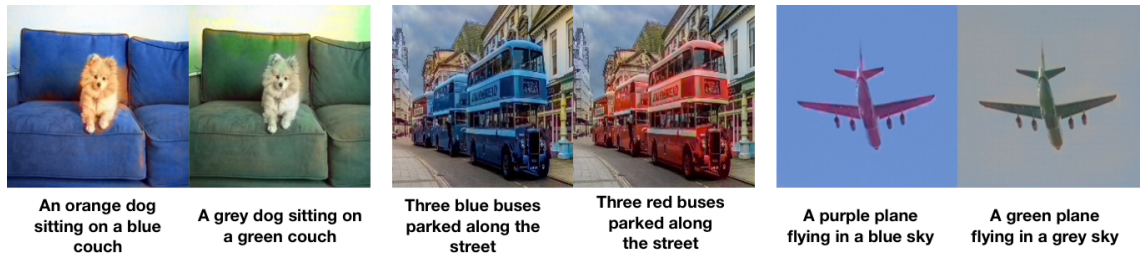

Figure 1: Three pairs of images whose colorizations are conditioned on corresponding captions by our FILM architecture. Our model can localize objects mentioned by the captions and properly color them.

## 2.2 Fully-convolutional networks for colorization

Following Zhang et al. (2016), we treat colorization as a classification problem in CIE *Lab* space: given only the lightness channel *L* of an image (i.e., a greyscale version), a fully-convolutional network predicts values for the two color channels $a$ and $b$. For efficiency, we deviate from Zhang et al. (2016) by quantizing the color channels into a 25×25 grid, which results in 625 labels for classification. To further speed up training, we use a one-hot encoding for the *ab* channels instead of soft targets as in Zhang et al. (2016); preliminary experiments showed no qualitative difference in colorization quality with one-hot targets. The contribution of each label to the loss is downweighted by a factor inversely proportional to its frequency in the training set, which prevents desaturated *ab* values. Our baseline network architecture (FCNN) consists of eight convolutional blocks, each of which contains multiple convolutional layers followed by batch normalization (Ioffe and Szegedy, 2015).[1] Next, we propose two ways to integrate additional *text*

---
[1] See Zhang et al. (2016) for complete architectural details. Code and pretrained models are available at https://github.com/superhans/colorfromlanguage.

| | *ab* Accuracy | | Human Experiments | | |
|---|---|---|---|---|---|
| **Model** | acc@1 | acc@5 | plaus. | qual. | manip. |
| FCNN | 15.4 | 45.8 | 20.4 | 32.6 | N/A |
| CONCAT | 17.9 | 50.3 | 39.0 | **34.1** | 77.4 |
| FILM | **23.7** | **60.5** | **40.6** | 32.1 | **81.2** |

Table 1: While FILM is the most accurate model in *ab* space, its outputs are about as contextually plausible as CONCAT's according to our *plausibility* task, which asks workers to choose which model's output best depicts a given caption (however, both models significantly outperform the language-agnostic FCNN). This additional plausibility does not degrade the output, as shown by our *quality* task, which asks workers to distinguish an automatically-colorized image from a real one. Finally, our caption *manipulation* experiment, in which workers are guided by a caption to select one of three outputs generated with varying color words, shows that modifying the caption significantly affects the outputs of CONCAT and FILM.

input into FCNN.

## 2.3 Colorization conditioned on language

Given an image *I* paired with a unit of text *T*, we first encode *T* into a continuous representation $h$ using the last hidden state of a bi-directional LSTM (Hochreiter and Schmidhuber, 1997). We integrate $h$ into every convolutional block of the FCNN, allowing language to influence the computation of all intermediate feature maps.

Specifically, say $\mathbf{Z}_n$ is the feature map of the $n$th convolutional block. A conceptually simple way to incorporate language into this feature map is to concatenate $h$ to the channels at each spatial location $i, j$ in $\mathbf{Z}_n$, forming a new feature map

$$\mathbf{Z}'_{n_{i,j}} = [\mathbf{Z}_{n_{i,j}}; h]. \qquad (1)$$

While this method of integrating language with images (**CONCAT**) has been successfully used for other vision and language tasks (Reed et al., 2016; Feichtenhofer et al., 2016), it requires considerably more parameters than the **FCNN** due to the additional language channels.

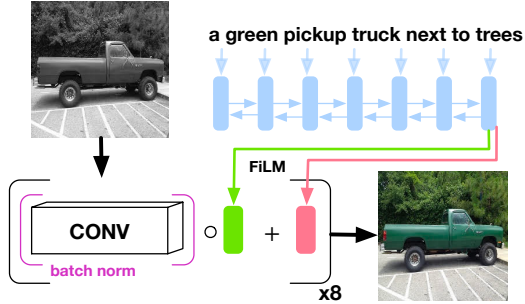

Figure 2: **FILM** applies feature-wise affine transformations (conditioned on language) to the output of each convolutional block in our architecture.

Inspired by recent work on visual question answering, we also experiment with a less parameter-hungry approach, feature-wise linear modulation (Perez et al., 2018, **FILM**), to fuse the language and visual representations. Since the activations of **FILM** layers have attention-like properties when trained on **VQA**, we also might expect **FILM** to be better at localizing objects from language than **CONCAT** on colorization (see Figure 4 for heatmap visualizations).

**FILM** applies a feature-wise affine transformation to the output of each convolutional block, where the transformation weights are conditioned on language (Figure 2). Given $\mathbf{Z}_n$ and $\mathbf{h}$, we first compute two vectors $\boldsymbol{\gamma}_n$ and $\boldsymbol{\beta}_n$ through linear projection,

$$\boldsymbol{\gamma}_n = \mathbf{W}_{n\gamma}\mathbf{h} \qquad \boldsymbol{\beta}_n = \mathbf{W}_{n\beta}\mathbf{h}, \qquad (2)$$

where $\mathbf{W}_{n\gamma}$ and $\mathbf{W}_{n\beta}$ are learned weight matrices. The modulated feature map then becomes

$$\mathbf{Z}'_{n_{i,j}} = (1 + \boldsymbol{\gamma}_n) \circ \mathbf{Z}_{n_{i,j}} + \boldsymbol{\beta}_n, \qquad (3)$$

where ∘ denotes the element-wise product. Compared to **CONCAT**, **FILM** is parameter-efficient, requiring just two additional weight matrices per feature map.

## 3 Experiments

We evaluate **FCNN**, **CONCAT**, and **FILM** using accuracy in *ab* space (shown by Zhang et al. (2016) to be a poor substitute for plausibility) and with crowdsourced experiments that ask workers to judge colorization *plausibility*, *quality*, and the colorization flexibly reflects language *manipulations*. Table 1 summarizes our results; while there is no clear winner between **FILM** and **CONCAT**, both rely on language to produce higher-quality colorizations than those generated by **FCNN**.

### 3.1 Experimental setup

We train all of our models on the 82,783 images in the MSCOCO (Lin et al., 2014) training set, each of which is paired with five crowdsourced captions. Training from scratch on MSCOCO results in poor quality colorizations due to a combination of not enough data and increased image complexity compared to ImageNet (Russakovsky et al., 2015). Thus, for our final models, we initialize all convolutional layers with a **FCNN** pretrained on ImageNet; we finetune both **FILM** and **CONCAT**'s convolutional weights during training. To automatically evaluate the models, we compute top-1 and top-5 accuracy in our quantized *ab* output space[2] on the MSCOCO validation set. While **FILM** achieves the highest *ab* accuracy, **FILM** and **CONCAT** do not significantly differ on crowdsourced evaluation metrics.

### 3.2 Human experiments

We run three human evaluations of our models on the Crowdflower platform to evaluate their plausibility, overall quality, and how well they condition their output on language. Each evaluation is run using a random subset of 100 caption/image pairs from the MSCOCO validation set,[3] and we obtain five judgments per pair.

*Plausibility* given caption: We show workers a caption along with three images generated by **FCNN**, **CONCAT**, and **FILM**. They choose the image that best depicts the caption; if multiple images accurately depict the caption, we ask them to choose the most realistic. **FCNN** does not receive the caption as input, so it makes sense that its output is only chosen 20% of the time; there is no significant difference between **CONCAT** and **FILM** in plausibility given the caption.

---
[2]We evaluate accuracy at the downsampled 56×56 resolution at which our network predicts colorizations. For human experiments, the prediction is upsampled to 224×224.
[3]We only evaluate on captions that contain one of ten "color" words (e.g., red, blue purple).

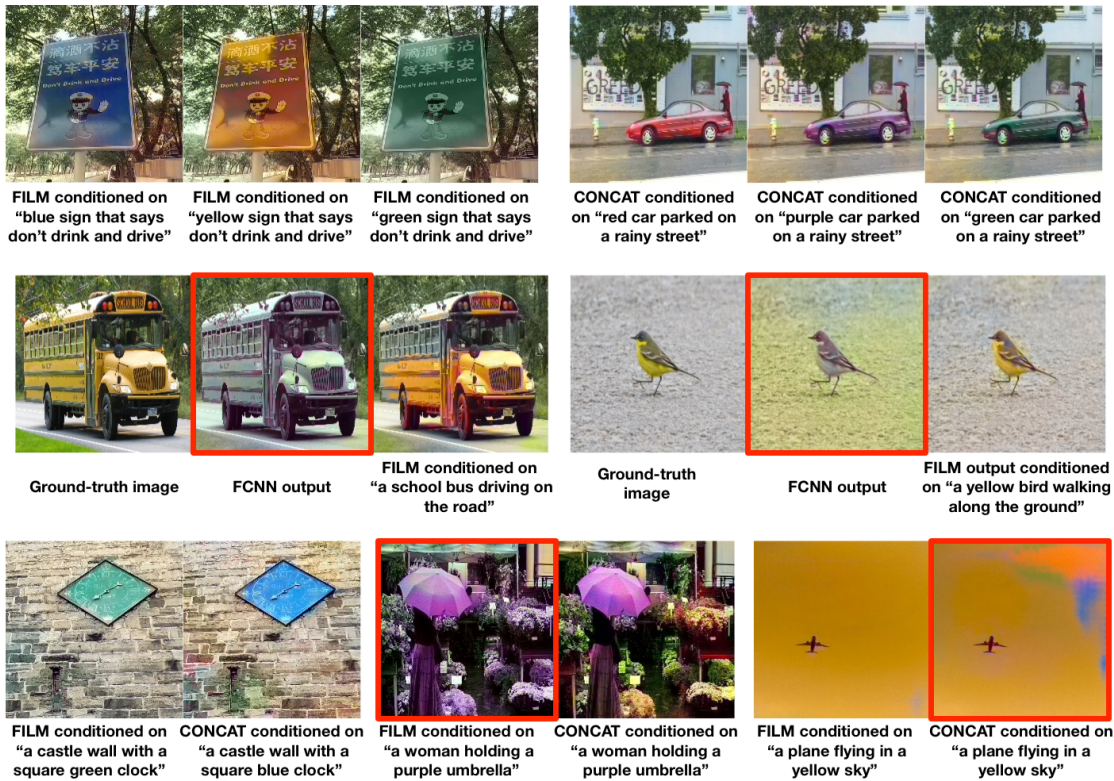

Figure 3: The top row contains successes from our caption manipulation task generated by FILM and CONCAT, respectively. The second row shows examples of how captions guide FILM to produce more accurate colorizations than FCNN (failure cases outlined in red). The final row contains, from left to right, particularly eye-catching colorizations from both CONCAT and FILM, a case where FILM fails to localize properly, and an image whose unnatural caption causes artifacts in CONCAT.

**Colorization *quality*:** Workers receive a pair of images, a ground-truth MSCOCO image and a generated output from one of our three architectures, and are asked to choose the image that was *not* colored by a computer. The goal is to fool workers into selecting the generated images; the "fooling rates" for all three architectures are comparable, which indicates that we do not reduce colorization quality by conditioning on language.

**Caption *manipulation*:** Our last evaluation measures how much influence the caption has on the CONCAT and FILM models. We generate three different colorizations of a single image by swapping out different colors in the caption (e.g., *blue car*, *red car*, *green car*). Then, we provide workers with a single caption (e.g., *green car*) and ask them to choose which image best depicts the caption. If our models cannot localize and color the appropriate object, workers will be unable to select an appropriate image. Fortunately, CONCAT and FILM are both robust to caption manipulations (Table 1).

## 4 Discussion

Both CONCAT and FILM can manipulate image color from captions (further supported by the top row of Figure 3). Here, we qualitatively examine model outputs and identify potential directions for improvement.

Language-conditioned colorization depends on correspondences between language and color statistics (*stop signs* are always red, and *school buses* are always yellow). While this extra information helps us produce more plausible colorizations compared to language-agnostic models (second row of Figure 3), it biases models trained on natural images against unnatural colorizations. For example, the yellow sky produced by CONCAT in the bottom right of Figure 3 contains blue artifacts because skies are usually blue in MSCOCO. Additionally, our models are limited by the lightness channel $L$ of the greyscale image, which prevents dramatic color shifts like black-to-white. Smaller objects are also problematic; often, colors will "leak" into smaller objects from larger ones, as shown by FILM's colorizations of purple plants (Figure 3, bottom-middle) and yellow tires (middle-left).

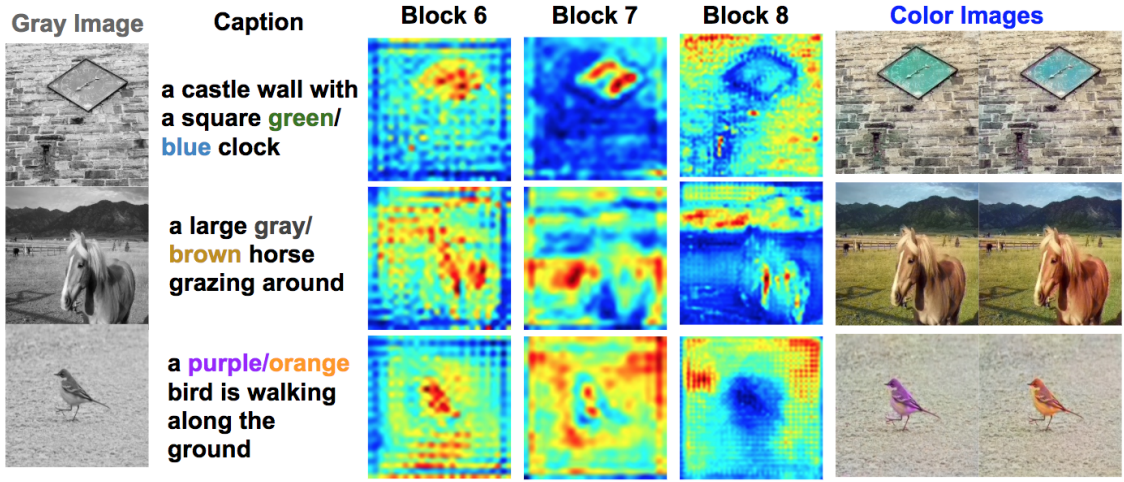

Figure 4: Examples of intermediate layer activations while generating colorized images using the FILM network. These activation maps correspond to the mean activation immediately after the FILM layers of the sixth, seventh, and eighth blocks. Interestingly, the activations after the FILM layer of Block 6 always seems to focus on the object that is to be colorized, while those of Block 8 focus almost exclusively on the background. The activation maps do not significantly differ when color words in the caption are manipulated; therefore, we show maps only for the first color word in these examples.

Figure 4 shows activation maps from intermediate layers generated while colorizing images using the FILM network. Each intermediate layer is captured immediately after the FILM layer and is of dimension $h \times w \times c$ (e.g., $112 \times 112 \times 64$, $28 \times 28 \times 512$, etc.), where $h$ is the height of the feature map, $w$ is its width, and $c$ is the number of channels.[4] On inspection, the first few activation maps correspond to edges and are not visually interesting. However, we notice that the sixth activation map usually focuses on the principal subject of the image (such as a car or a horse), while the eighth activation map focused everywhere but on that subject (i.e., entirely on the background). This analysis demonstrates that the FILM layer emulates visual attention, reinforcing similar observations on visual QA datasets by Perez et al. (2018).

## 5 Future Work

While these experiments are promising, that there are many avenues to improve language-conditioned colorization. From a vision perspective, we would like to more accurately colorize parts of objects (e.g., a person's shoes); moving to more complex architectures such as variational autoencoders (Deshpande et al., 2017) or PixelCNNs (Guadarrama et al., 2017) might help here, as could increasing training image resolution. We also plan on using refinement networks (Shrivastava et al., 2017) to correct for artifacts in the colorized output image. On the language side, moving from explicitly specified colors to abstract or emotional language is a particularly interesting. We plan to train our models on dialogue/image pairs from datasets such as COMICS (Iyyer et al., 2017) and visual storytelling (Huang et al., 2016); these models could also help learn powerful joint representations of vision and language to improve performance on downstream prediction tasks.

## Acknowledgement

Manjunatha and Davis are partially supported by the Office of Naval Research under Grant N000141612713: Visual Common Sense Reasoning. Boyd-Graber is supported by NSF Grant IIS-1652666. We thank Richard Zhang for helpful comments on our ideas to make training the colorization more efficient.

---

[4] We compute the mean across the $c$ dimension and scale the resulting $h \times w$ feature map between the limits [0, 255].